\documentclass[conference,10pt]{IEEEtran}

\usepackage{multicol}
\usepackage{amsmath,epsfig}
\usepackage{amssymb}
\usepackage{booktabs}
\usepackage{amsthm}
\usepackage{graphicx}
\usepackage{caption}
\usepackage{subcaption}
\usepackage{rotating}
\usepackage{floatflt} 
\usepackage{paralist} 
\usepackage{algorithm} 
\usepackage{xcolor}
\usepackage{color}
\usepackage{epstopdf}
\usepackage[normalem]{ulem}
\usepackage{float}
\usepackage{todonotes}
\usepackage{comment}
\usepackage{todonotes}

\theoremstyle{definition}

\newtheorem{myLem}{Lemma}

\newtheorem{myThm}{Theorem}
\newtheorem{myCorollary}{Corollary}
\newcommand{\R}{\ensuremath{\mathbb{R}}}

\newcommand{\argmin}{\mathbf{arg min}}

\def\x{\mathbf{x}}

\def\y{\mathbf{y}}

\def\N{\mathcal{N}}

\def\M{\mathcal{M}}
\def\U{\mathcal{U}}

\DeclareMathOperator{\Dd}{D}

\DeclareMathOperator{\Hm}{H}

\DeclareMathOperator{\X}{X}
\DeclareMathOperator{\W}{W}

\IEEEoverridecommandlockouts
\begin{document}

\title{A statistical perspective of  sampling scores for linear regression}
\author{Siheng Chen$^{1,2}$, Rohan Varma$^1$, Aarti Singh$^2$, Jelena Kova\v{c}evi\'c$^{1,3}$ \thanks{The authors gratefully acknowledge support from the NSF through awards 1130616, 1421919, IIS-1247658, IIS-1252412 and AFOSR YIP FA9550-14-1-0285. Due to the lack of space, the full proofs of results appear in~\cite{ChenVSK:15g}.
}  \\
$^1$ Dept. of ECE, $^2$ Dept. of Machine Learning, $^3$Dept. of BME, \\
  Carnegie Mellon University,  Pittsburgh, PA, USA}	
\date{}									
\maketitle


\begin{abstract}
In this paper, we consider a statistical problem of learning a linear model from noisy samples.  Existing work has focused on approximating the least squares solution by using leverage-based scores as an importance sampling distribution. However, no finite sample statistical guarantees and no computationally efficient optimal sampling strategies have been proposed. To evaluate the statistical properties of different sampling strategies, we propose a simple yet effective estimator, which is easy for theoretical analysis and is useful in multitask linear regression. We derive the exact mean square error of the proposed estimator for any given sampling scores. Based on minimizing the mean square error, we propose the optimal sampling scores for both estimator and predictor, and show that they are influenced by the noise-to-signal ratio. Numerical simulations match the theoretical analysis well.
\end{abstract}

\section{Introduction}
In many engineering problems, it is expensive to measure and store a large set of samples. Motivated by this, there has been a great deal of work on developing an effective sampling strategy for a variety of matrix-based problems, including compressed sensing~\cite{Candes:06, Donoho:06}, adaptive sampling for signal recovery and detection~\cite{TroppLDRB:10, HauptCN:11, DavenportMNW:15}, adaptive sampling for matrix approximation and completion~\cite{KrishnamurthyS:13, PatelGDMB:15,WangS:15a} and many others. At the same time, motivated by computational efficiency, statistically effective sampling strategies are developed for least-squares approximation~\cite{DrineasMM:06}, least absolute deviations regression~\cite{ClarksonDMMMW:13} and low-rank matrix approximation~\cite{MahoneyD:09}. 

One way to develop a sampling strategy is to design a data-dependent importance sampling distribution from which to sample the data. A widely used sampling distribution to select rows and columns of the input matrix are the leverage scores, defined formally in Section 2. Typically, the leverage scores are computed approximately~\cite{DrineasMMW:12,ClarksonDMMMW:13}, or otherwise, a random projection is used to precondition by approximately
uniformizing them~\cite{AvronMT:10,AilonC:10, DrineasMMS:11,ClarksonDMMMW:13,MengSM:14}. A detailed discussion of this approach can be found in the recent review on randomized algorithms for matrices and matrix-based data problems~\cite{Mahoney:11}. Even though leverage-based sampling distributions are widely used, for many problems, it is unclear that why they might work well from a statistical perspective.

For the problem of low-rank matrix approximation,~\cite{WangS:15b} provided an extensive empirical evaluation of several sampling distributions and showed that iterative norm sampling outperforms leverage score based sampling empirically.~\cite{YangZJZ:15} showed that for low-rank matrix approximation, the square root of the leverage score based sampling statistically outperforms uniform sampling all the time and outperforms leverage score based sampling when the leverage scores are nonuniform. Further, the authors proposed a constrained optimization problem resulting in optimal sampling probabilities, which are between leverage scores and their square roots.

For the problem of linear regression, the optimal sampling strategies for estimation and prediction are known to be obtained via $A$ and $G$-optimality criteria, respectively~\cite{Pukelsheim:06}, but the resulting sampling strategies are combinatorial. The computationally efficient optimal sampling strategies are still unclear.~\cite{MaMY:15} showed that based on the sampled least squares, neither leverage score based sampling nor uniform sampling outperforms the other. The leveraging based estimator could suffer from a large
variance when the leverage scores are nonuniform while uniform sampling is less vulnerable to small leverage scores.~\cite{ZhuMMY:15} only analyzed the asymptotic behavior of the sampled least squares and the corresponding optimal sampling is still unclear.

In this paper, we propose an estimator with computationally efficient sampling strategy that also comes with closed form and finite sample guarantees on performance. We derive the exact mean square error (MSE) of this estimator and show the optimal sampling distribution for both estimator and predictor. The results of the statistical analysis can be summarized as follows: (1) the proposed estimator is unbiased for any sampling distribution; (2) the optimal sampling distribution is influenced by the noise-to-signal ratio; (3) the optimal sampling distribution involves a tradeoff between the leverage scores and the square root of the leverage scores. We further provide an empirical evaluation of the proposed algorithm on synthetic data under various settings. This empirical evaluation clearly shows the efficacy of the derived optimal sampling scores for sampling large-scale data in order to learn a linear model from the noisy samples.

Except for the simple statistical analysis, the proposed sampled projection estimator is useful for the multitask linear regression. This is because the computational cost is less when we estimate multiple coefficient vectors based on the same data matrix. This is crucial for many applications on sensor networks and electrical systems~\cite{ChenVSK:15a}. For example, traffic speeds are time-evolving data supported on an urban traffic network. Based on the same urban traffic network, it is potentially efficient to use the proposed estimator to estimate the traffic speeds at all the intersections from a small number of samples every period of time. The proposed estimator also can be applied to estimate corrupted measurements in sensor networks~\cite{ChenSMK:14}, uncertain voltage states in power grids~\cite{WengNI:13} and many others.

The main contributions are (1) we derive the exact MSE of a natural modification of the least squares estimator; and (2) we derive the optimal sampling distributions for the estimator and the predictor, and show that they are influenced by the noise-to-signal ratio and they are related to leverage scores.

\section{Background}
We consider a simple linear model
\begin{eqnarray*}
	\y  \ = \ \X \beta^{(0)} + \epsilon,
\end{eqnarray*}
where $\y \in \R^n$ is a response vector, $\X \in \R^{n \times p}$ is a data matrix ($n > p$), $\beta^{(0)} \in \R^p$ is a coefficient vector, and $\epsilon  \in \R^n$ is a zero mean noise vector with independent and identically distributed entries, and covariance matrix $\sigma^2 I$. The task is, given $\X$, $\y$, we aim to estimate $\beta^{(0)}$. In this case, the unknown coefficient vector $\beta^{(0)}$ is usually estimated via the least squares as
\begin{eqnarray}
\label{eq:ls}
	\widehat{\beta}  \ = \   \arg \min_{\beta}  \left\| \y - \X \beta \right\|_2^2 \ = \  (\X^T \X)^{-1} \X^T \y \ = \ \X^{\dagger} \y
\end{eqnarray}
where $\X^{\dagger} =  (\X^T \X)^{-1} \X^T \in \R^{p \times n}$. The predicted response vector is then $\widehat{\y}  = \Hm \y$, where $\Hm = \X \X^{\dagger} $ is the hat matrix. The diagonal elements of $\Hm$ are the leverage scores, that is, $\Hm_{i, i}$ is the leverage of the $i$th sample. The statistical leverage scores are widely used for detecting outliers and influential data~\cite{HoaglinW:78,VellemanW:81,DrineasMMW:12}.

In some applications, it is expensive to sample the entire response vector. Some previous works consider sampling algorithms computing the approximate solutions to the overconstrained least squares problem~\cite{DrineasMM:06, Mahoney:11, DhillonLFU:13, MaMY:15}. These algorithms choose a small number of rows of $\X$ and the corresponding elements of $\y$, and use least squares on the samples to estimate $\beta^{(0)}$. Formally, let $\M = (\M_1, \cdots, \M_{m})$ be the sequence of sampled indices, called~\emph{sampling set}, with $|\M| = m$ and $\M_i \in \{1, \cdots, n\}$. A sampling matrix $\Psi$ is a linear mapping from
$\R^n$ to $\R^m$ that describes the process of sampling with replacement,
\begin{equation*}
\label{eq:Psi}
 \Psi_{i,j} = 
  \left\{ 
    \begin{array}{rl}
      1, & j = \M_i;\\
      0, & \mbox{otherwise}.
  \end{array} \right.
\end{equation*}
When the $j$th element of $\y$ is chosen in the $i$th random trial, then $\M_i = j$ and $\Psi_{i,j} = 1$. The goal is {\bf given $\X$, we design the sampling operator $\Psi$ to obtain samples $\Psi \y$, and then, estimate $\beta^{(0)}$}. Here we focus on experimental design of sampling operator, that is, the sampling strategy needs to be designed in advance of collecting any $y_i$.

The choice of samples is an important degree of freedom when studying the corresponding quality of approximation. In this paper, we employ random sampling with an underlying probability distribution to choose samples. Let $\{ \pi_i \}_i^n$ be a probability distribution, where $\pi_i$ denotes the probability to choose the $i$th sample in each random trial. We consider two choices of the probability distribution:~\emph{uniform sampling} means that the sample indices are chosen from $\{1, 2, \cdots, n\}$ independently and randomly; and~\emph{sampling score-based sampling} which means that the sample indices are chosen from an importance sampling distribution that is proportional to a sampling score that is computed from the data matrix\footnote{the terms of sampling distribution and sampling scores mean the same thing in this paper}. A widely-used sampling score is the leverage scores of the data matrix. Given the samples, one way to estimate $\beta^{(0)}$ is by solving a weighted least square problem:
\begin{eqnarray}
\label{eq:SampleLS}
	\widehat{\beta}  & = &   \arg \min_{\beta}  \left\| \Dd \Psi \y -  \Dd \Psi \X \beta \right\|_2^2
	\nonumber
	\\
	& = &  \left( \Dd \Psi \X \right)^{\dagger}  \Dd \Psi \y,
\end{eqnarray}
where $\Dd \in \R^{m \times m}$ is a diagonal rescaling matrix with $\Dd_{i,i} = 1/\sqrt{m \pi_j}$ when $\Psi_{i,j} = 1$. This is called~\emph{sampled least squares} (SampleLS)~\cite{MaMY:15}. The computational cost of taking pseudo-inverse of $\Dd \Psi \X \in \R^{m \times p}$ is $O(2p^2m + p^3)$. The inverse term is involved with the sampling operator, but the computation is much cheaper than taking pseudo-inverse of $\X \in \R^{n \times p}$, which takes $O(2p^2n + p^3)$ ($n \geq m \geq p$). However, it is hard to show the exact MSEs and the optimal sampling scores of SampleLS.~\cite{MaMY:15} shows that, based on SampleLS, neither leverage score based sampling nor uniform sampling dominates the other from from a statistical perspective.

\section{Proposed Method}
 To have a deeper statistical understanding of this task, we propose a similar, but simpler algorithm to deal with the same task, but it is easy to show its corresponding exact MSEs and the corresponding optimal sampling scores.

\subsection{Problem Formulation}
Similarly to~\eqref{eq:SampleLS}, we estimate $\beta^{(0)}$ by solving the following problem:
\begin{eqnarray}
\label{eq:proj}
\widehat{ \beta } & = &  \argmin_{\beta}  \left\|  \Psi^T \Psi \Dd^2 \Psi^T  \Psi \y - \X \beta \right\|_2^2 
\nonumber
\\
& = &  \X^{\dagger} \Psi^T \Psi \Dd^2 \Psi^T  \Psi \y,
\end{eqnarray}
where  $\Dd \in \R^{n \times n}$ is the same diagonal rescaling matrix in~\eqref{eq:SampleLS}. We call~\eqref{eq:proj}~\emph{sampled projection} (SampleProj). This is akin to zero-filling the unobserved entries of $\y$ and rescaling the non-zero entries to make it unbiased. Comparing to the ordinary least squares, SampleProj does not benefit the computational efficiency because it still needs to taking pseudo-inverse of $\X \in \R^{n \times p}$ and the computation involves the factor $n$; however, it is still useful when taking samples is expense. Comparing to SampleLS, this algorithm is more appealing when we aim to estimate multiple $\beta^{(0)}$s based on the same data matrix $\X$ because the inverse term is not involved with the sampling operator.

\subsection{Statistical Analysis}
We obtain the exact mean squared error and an optimal sampling distribution for SampleProj. Our main contributions based on SampleProj can be summarized as follows:
\begin{itemize}
\item the estimator is unbiased for any sampling scores (Lemma~\ref{lem:unbias} );
\item the closed-form solution (finite sample) on the MSE of the estimator and the predictor (Theorem~\ref{thm:MSE}); and
\item analytic optimal sampling scores of the estimator and predictor are provided (Theorem~\ref{thm:optimal}).
\end{itemize}

Due to the lack of space, the full proofs of results appear in~\cite{ChenVSK:15g}

\begin{myLem} 
\label{lem:unbias}
The estimator $\widehat{ \beta}$ in SampleProj is an unbiased estimator of $\beta^{(0)}$, that is, 
$
	\mathbb{E} \left[  \widehat{ \beta }  \right] \ = \ \beta^{(0)}.
$
\end{myLem}

\begin{myLem} 
\label{lem:var}
The covariance of the estimator $\widehat{ \beta}$ in SampleProj has the following property:
\begin{eqnarray*}
&&	{\rm  Tr} ( {\rm Covar}  \left[ \widehat{ \beta}    \right] ) 
	\ = \  \mathbb{E}  \left\| \widehat{ \beta} - \mathbb{E}   \left[ \widehat{ \beta} \right]   \right\|^2 
	\\
	& = &   \sum_{i=1}^{p}  \left( \sum_{l=1}^n  \frac{1}{ m \pi_l }  
(\X^{\dagger})_{i,l}^2 \left( ( \X \beta^{(0)} )_l^2 +  \sigma^2 \right) - \frac{1}{m} ( \beta^{(0)})_i^2 \right),
\end{eqnarray*}
where $ \sigma^2$ is the variance of the Gaussian random noise.
\end{myLem}

Combining the results of Lemmas~\ref{lem:unbias} and~\ref{lem:var}, we obtain the exact MSE of both the estimator and the predictor.
\begin{myThm}
\label{thm:MSE}
Let $\widehat{ \beta}$ be the solution of SampleProj with sampling distribution of $\{ \pi_i \}_{i=1}^n$.
The mean squared error of the estimator $\widehat{ \beta}$  is
\begin{eqnarray*}
  \mathbb{E}  \left\|  \widehat{ \beta}  - \beta^{(0)}  \right\| ^2 
 \ = \   
{\rm Tr} \left( \X^{\dagger} \W_{\rm C}  (\X^{\dagger})^T \right) - \frac{  1}{m} \left\|  \beta^{(0)} \right\|_2^2,
\end{eqnarray*}
where $(\W_{\rm C})_{l,l} =  \left( ( \X \beta^{(0)})_l^2 +  \sigma^2 \right)/( m  \pi_l )$. The mean squared error of the predictor $\X \widehat{ \beta}$   is
\begin{eqnarray*}
 && \mathbb{E}  \left\|  \X \widehat{ \beta}  - \X \beta^{(0)}  \right\| ^2  
 \\
 & = &    
{\rm Tr} \left( \Hm \W_{\rm C} \right) -  \frac{1}{m} \left\|  \X \beta^{(0)} \right\|_2^2.
\end{eqnarray*}
\end{myThm}

We next optimize over the mean squared errors and obtain the optimal sampling scores.
\begin{myThm}
\label{thm:optimal}
The optimal sampling score to minimize the mean squared error of the estimator $\widehat{ \beta}$ is 
\begin{eqnarray*}
 \pi_l \propto \sqrt{ \left( \X (\X^T \X)^{-2} \X^T   \right)_{l,l}  \left( (\X \beta^{(0)})_l^2  + \sigma^2 \right) };
\end{eqnarray*}
The optimal sampling scores to minimize the MSE of the predictor $\X \widehat{ \beta}$ is 
\begin{eqnarray*}
 \pi_l \propto \sqrt{ \Hm_{l,l} \left(  (\X \beta^{(0)})_l^2  + \sigma^2  \right) }.
\end{eqnarray*}
\end{myThm}

Theorem~\ref{thm:optimal} shows that the optimal sampling scores depend on the signal strength and the noise.  In practice, we cannot access to $\X \beta^{(0)}$ and we need to approximate  the ratio between each $(\X \beta^{(0)})_l$ and $\sigma^2$. For active sampling, we can collect the feedback to approximate each signal coefficient $(\X \beta^{(0)})_l$; for experimentally designed sampling, we approximate beforehand. One way is to use the Cauchy-Schwarz inequality to bound $x_i$,
\begin{eqnarray*}
(\X \beta^{(0)})_l  & = &  | \x_l^T \beta^{(0)} | 
\  \leq \  \left\|  \x_l \right\|_2  \left\| \beta^{(0)} \right\|_2.
\end{eqnarray*}
An upper bound of the MSE of the estimator is then
\begin{eqnarray*}
  \mathbb{E}  \left\|  \widehat{ \beta}  - \beta^{(0)}  \right\| ^2 
 \ \leq \   
{\rm Tr} \left( \X^{\dagger} \W_{\rm \bar{C}}  (\X^{\dagger})^T \right) - \frac{  1}{m} \left\|  \beta^{(0)} \right\|_2^2,
\end{eqnarray*}
where $(\W_{\rm \bar{C}})_{l,l} =  \left( \left\|  \x_l \right\|_2^2  \left\| \beta^{(0)} \right\|_2^2 \right)/( m  \pi_l )$. The corresponding sampling scores are 
 \begin{eqnarray}
  \label{eq:oss_esti}
  \pi_l  & \propto &  \sqrt{ \left( (\X^{\dagger})^T \X^{\dagger} \right)_{l,l}  \left( \left\|  \x_l \right\|_2^2  \left\| \beta^{(0)} \right\|_2^2 + \sigma^2 \right) }
	\nonumber 
  \\
  \nonumber
  & = & \sqrt{ \left( (\X^{\dagger})^T \X^{\dagger} \right)_{l,l}    \left\| \beta^{(0)} \right\|_2^2  \left( \left\|  \x_l \right\|_2^2 + \frac{ \sigma^2 } { \left\| \beta^{(0)} \right\|_2^2 }  \right)  }
  \\
  & \propto & \sqrt{ \left( \X (\X^T \X)^{-2} \X^T \right)_{l,l}  \left( \left( \X\X^T \right)_{l,l}  + {\rm NSR} \right)  },
\end{eqnarray}
where ${\rm NSR} = \sigma^2/\left\| \beta^{(0)} \right\|_2^2$.

When the column vectors of $\X$ are orthonormal,
$\X^T \X$ is the identity matrix, the sampling scores~\eqref{eq:oss_esti} are between the leverage scores and the square root of the leverage scores. 

\begin{myCorollary}
\label{cor:lev_esti}
Let the column vectors of $\X$ are orthonormal.
When ${\rm NSR} \rightarrow +\infty$, the sampling scores~\eqref{eq:oss_esti} are the square root of the leverage scores, that is, $\pi_l = \sqrt{\Hm_{l,l}}$, for all $l$. When ${\rm NSR} \rightarrow 0$, the sampling scores~\eqref{eq:oss_esti}  are the leverage scores, that is, $\pi_l = \Hm_{l,l}$, for all $l$.
\end{myCorollary}

An upper bound of the MSE of the predictor is then
\begin{eqnarray*}
  \mathbb{E}  \left\| \X \widehat{ \beta}  - \X \beta^{(0)}  \right\| ^2 
 \ \leq \   
{\rm Tr} \left( \Hm \W_{\rm \bar{C}} \right) -  \frac{1}{m} \left\|  \X \beta^{(0)} \right\|_2^2,
\end{eqnarray*}
where $(\W_{\rm \bar{C}})_{l,l} =  \left( \left\|  \x_l \right\|_2^2  \left\| \beta^{(0)} \right\|_2^2 \right)/( m  \pi_l )$. The corresponding optimal sampling scores are 
 \begin{eqnarray}
 \label{eq:oss_pred}
  \pi_l  & \propto &  \sqrt{ \Hm_{l,l}  \left( \left\|  \x_l \right\|_2^2  \left\| \beta^{(0)} \right\|_2^2 + \sigma^2 \right) }
 \nonumber \\
  & \propto & \sqrt{ \Hm_{l,l}  \left(  \left( \X\X^T \right)_{l,l} + {\rm NSR} \right)  }.
\end{eqnarray}

When NSR goes to infinity, the optimal sampling scores are always the square root of the leverage scores for any $\X$. When the column vectors of $\X$ are orthonormal,
the sampling scores~\eqref{eq:oss_pred} are the same with~\eqref{eq:oss_esti} and are between the leverage scores and the square root of the leverage scores. 

\begin{myCorollary}
\label{cor:lev_pred}
When ${\rm NSR} \rightarrow +\infty$, the sampling scores~\eqref{eq:oss_pred} are the square root of the leverage scores, that is, $\pi_l = \sqrt{\Hm_{l,l}}$, for all $l$. When the column vectors of $\X$ are orthonormal and ${\rm NSR} \rightarrow 0$, the sampling scores~\eqref{eq:oss_pred}  are the leverage scores, that is, $\pi_l = \Hm_{l,l}$, for all $l$.
\end{myCorollary}

\section{Empirical Evaluation}
In this section, we validate the proposed algorithm and statistical analysis on synthetic data.

\subsection{Synthetic Dataset}
\label{sec:data}
We generate a $1000 \times 20$ data matrix $\X$ from multivariate t-distribution with 1 degree of freedom and covariance matrix $\Sigma$, whose elements are $\Sigma_{i,j} = 2 \times 0.5^{|i-j|}$.
The  leverage scores of this matrix are nonuniform. This is the same as $T_1$ in~\cite{MaMY:15}. We then generate a $20 \times 1$ coefficient vector $\beta^{0}$, whose elements are drawn from uniform distribution $\U(0, 1)$. We test both noiseless and noisy cases. In the noiseless case, a response vector is $\y = \X \beta^{0}$; and in the noisy case, a response vector is $\y = \X \beta^{0} + \epsilon$, where $\epsilon \sim  \N(0, \sigma^2)$. We vary $\sigma$ as $5, 25, 50, 75, 100$. The corresponding ratios between $\left\| \epsilon \right\|_2^2$ and $\left\| \y \right\|_2^2$ are around $0.7\%, 15\%, 40\%, 60\%, 72\%$ and the corresponding NSR, $\sigma^2/\left\| \beta^{(0)} \right\|_2 $, are around $4, 100, 400, 800, 1600$.

\subsection{Experimental Setup}
\label{sec:setup}
We consider two key questions as follows:
\begin{itemize}
\item do different sampling scores make a difference?
\item does noise make a difference?
\end{itemize}
To answer these questions, we test the proposed SampleProj with different sampling scores for both noiseless and noisy cases. 
For each test, we compare $5$ sampling scores, including uniform sampling as Uniform (blue), leverage scores as Lev (red), square root of the leverage scores as sql-Lev (orange), sampling scores for the estimator~\eqref{eq:oss_esti} as opt-Est (purple), and sampling scores for the predictor~\eqref{eq:oss_pred} as opt-Pred (green). For opt-Est and opt-Pred, we use the true noise-to-signal ratios.  All the results are averaged over 500 independent runs. 

We measure the quality of estimation as
\begin{equation*}
{\rm err_{Est}} \ = \ \frac{ \left\| \widehat{\beta} - \beta^{(0)} \right\|_2 }{ \left\| \beta^{(0)} \right\|_2 },
\end{equation*}
where $\beta^{(0)} $ is the ground truth and $\widehat{\beta}$ is an estimator, and 
we measure the quality of prediction as
\begin{equation*}
{\rm err_{Pred}} \ = \ \frac{ \left\| \X \widehat{\beta} - \X \beta^{(0)} \right\|_2 }{ \left\| \X \beta^{(0)} \right\|_2 }.
\end{equation*}

\subsection{Results}
\begin{figure}[htb]
  \begin{center}
    \begin{tabular}{cc}
\includegraphics[width=0.48\columnwidth]{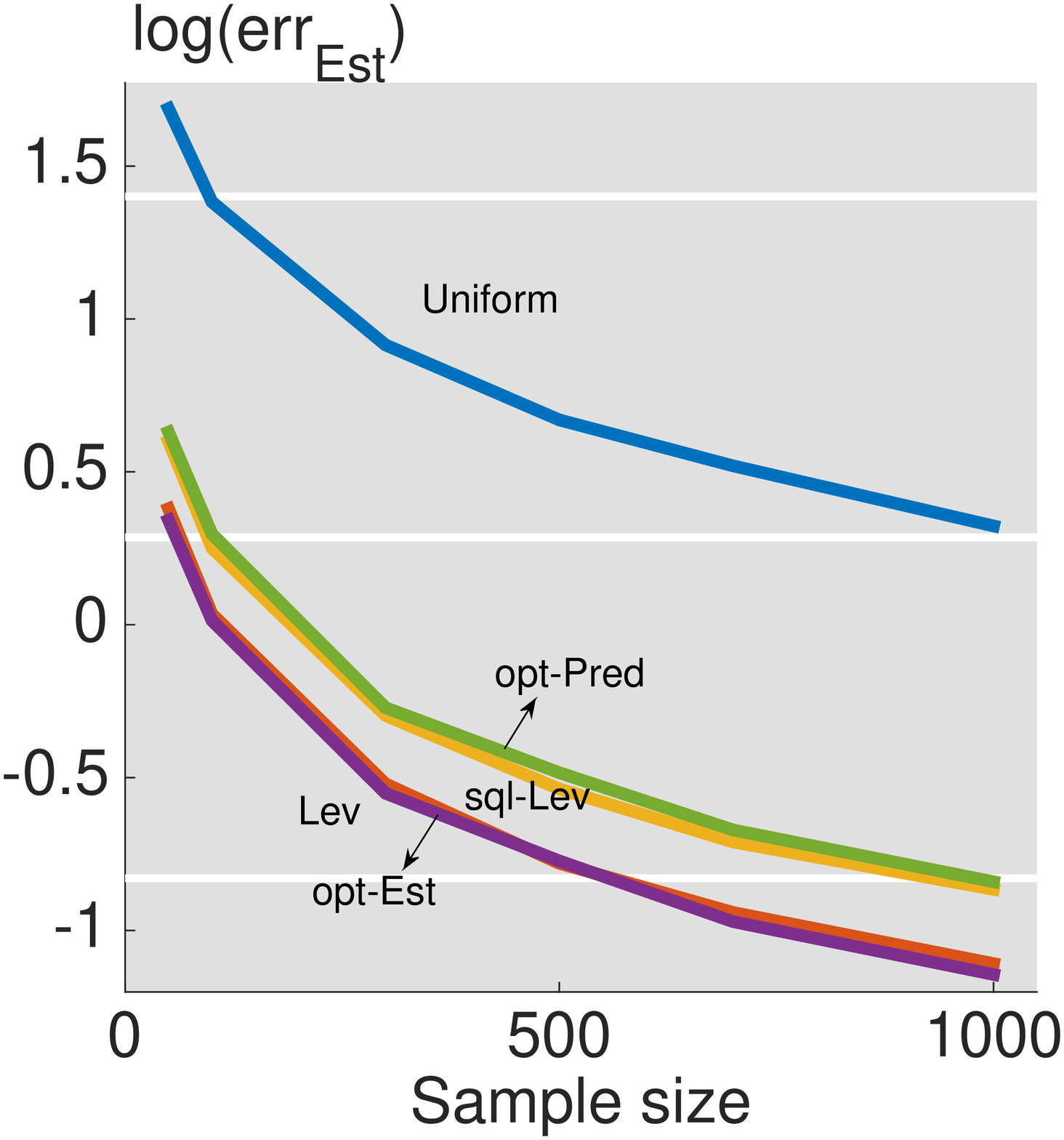}  & \includegraphics[width=0.48\columnwidth]{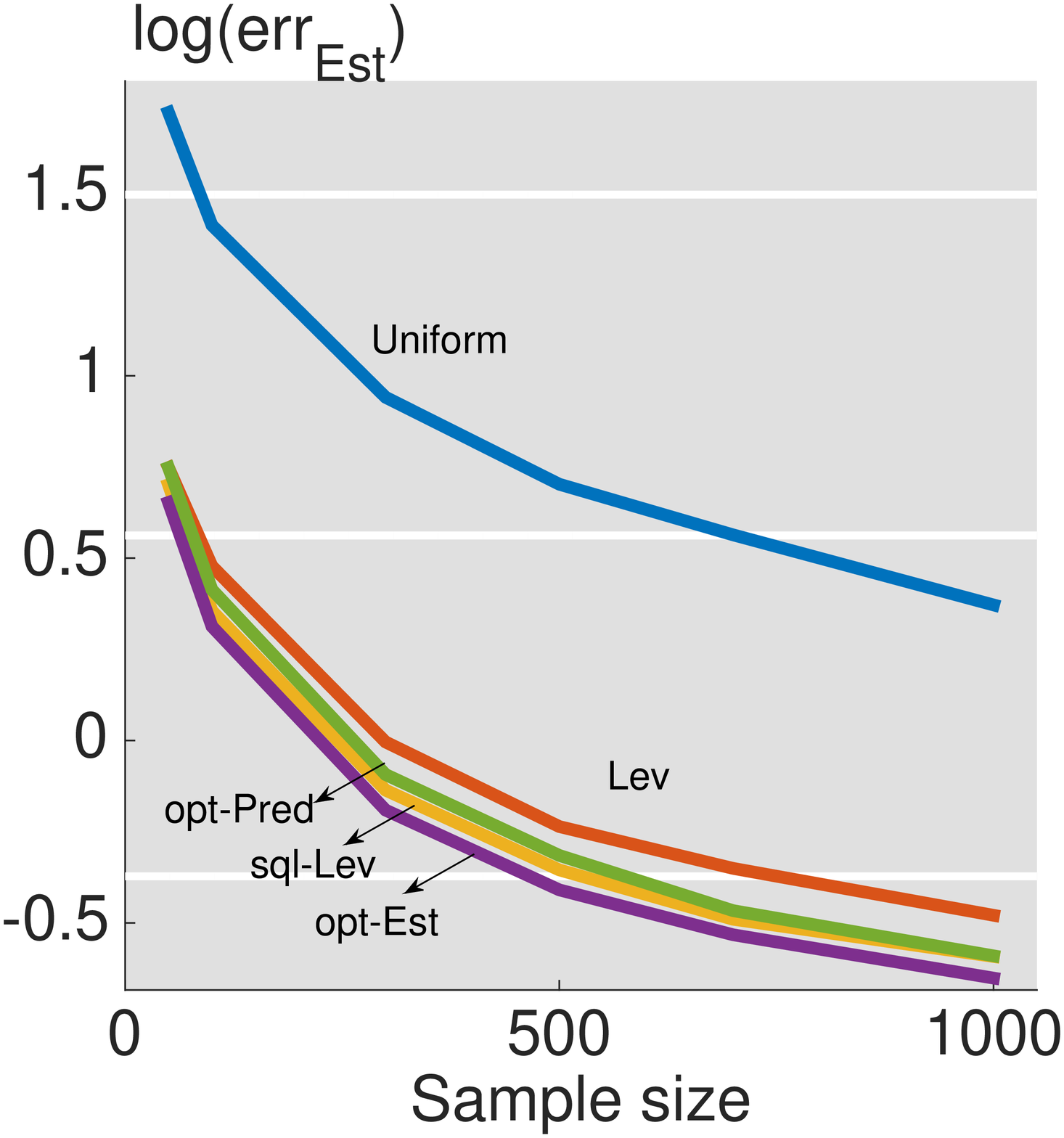} 
\\
 {\small (a) Noiseless.} & {\small (b) noisy $\sigma = 40$.} 
 \\ 
\end{tabular}
  \end{center}
   \caption{\label{fig:proj_t_t1_est} Comparison of estimation error for noisless and noisy cases of SampleProj in Task 1 as a function of sample size.}
\end{figure}

Figure~\ref{fig:proj_t_t1_est} compares the estimation error in both the noiseless and noisy scenarios of SampleProj as a function of sample size. We see that, as expected, the sampling scores for the estimator~\eqref{eq:oss_esti} outperform all other scores in both noiseless and noisy cases. Also, leverage scores perform well in the noiseless case, and square root of the leverage scores perform well in the noisy case, which matches the result in Corollary~\ref{cor:lev_esti}.

\begin{figure}[htb]
  \begin{center}
    \begin{tabular}{cc}
\includegraphics[width=0.48\columnwidth]{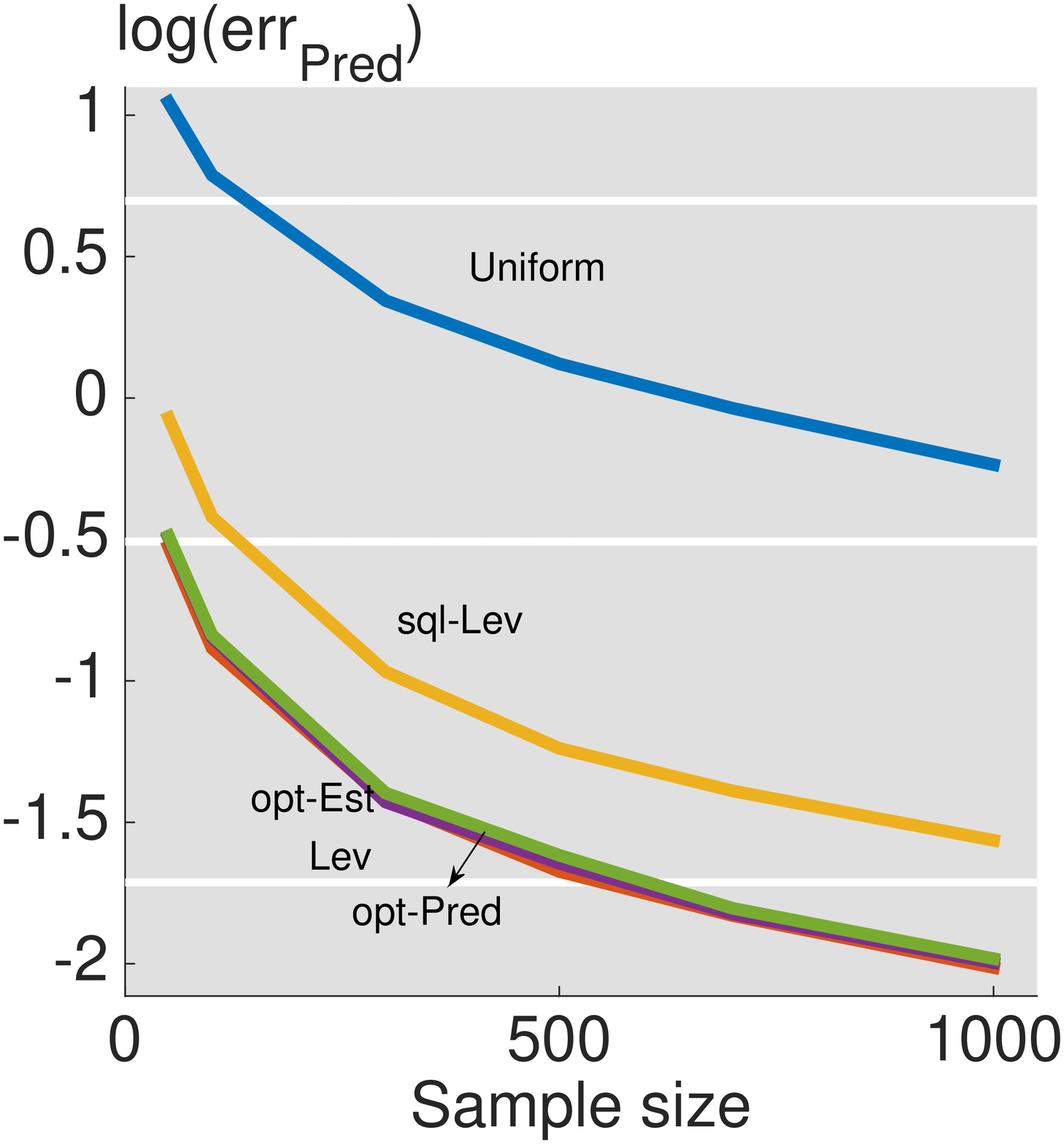}  & \includegraphics[width=0.48\columnwidth]{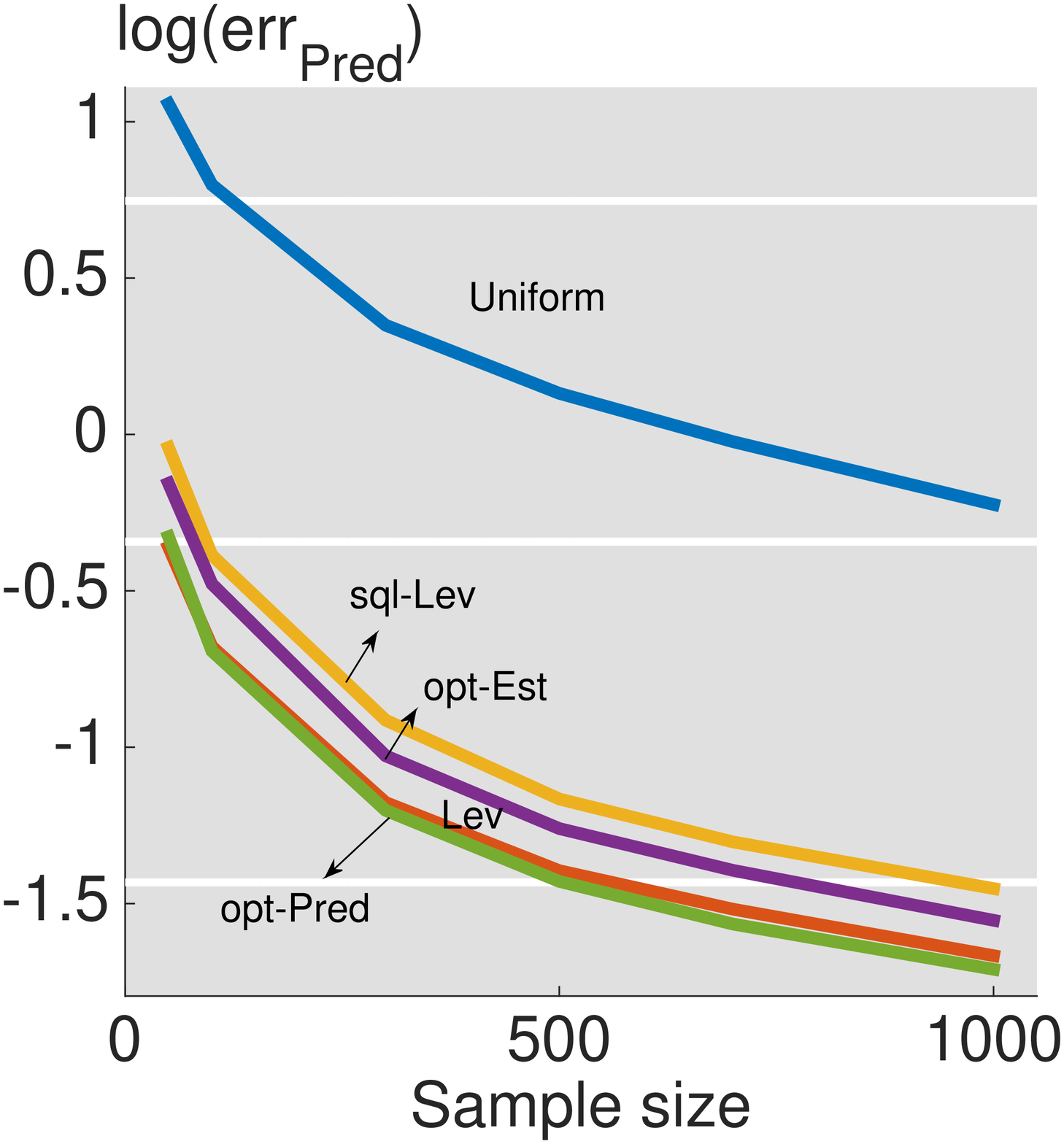} 
\\
 {\small (a) Noiseless.} & {\small (b) Noisy $\sigma = 40$.} 
 \\ 
\end{tabular}
  \end{center}
   \caption{\label{fig:proj_t_t1_pred} Comparison of prediction error for noisless and noisy cases of SampleProj as a function of sample size.}
\end{figure}

Figure~\ref{fig:proj_t_t1_pred} compares the prediction error for noiseless and noisy cases of SampleProj as a function of sample size. We see that, the sampling scores for the predictor~\eqref{eq:oss_pred} result in the best performance in both the noiseless and noisy cases; leverage scores perform better than square root of the leverage scores in the noiseless cases. In Corollary~\ref{cor:lev_esti}, we see that square root of the leverage scores should perform better than leverage scores in a high noise case, thus, we suspect that the noise level is not high enough to see the trend.

\begin{figure}[htb]
  \begin{center}
    \begin{tabular}{cc}
\includegraphics[width=0.48\columnwidth]{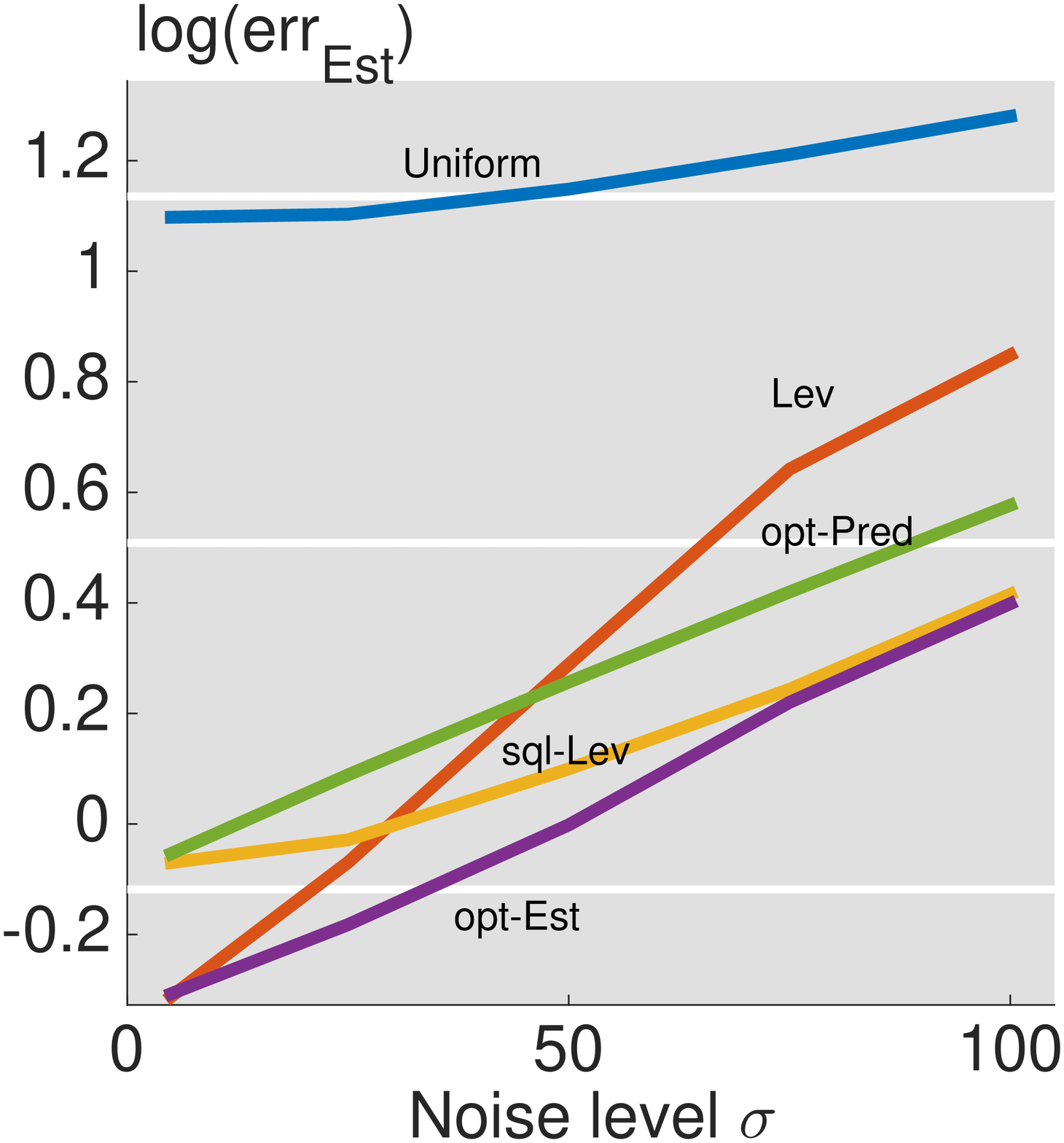}  & \includegraphics[width=0.48\columnwidth]{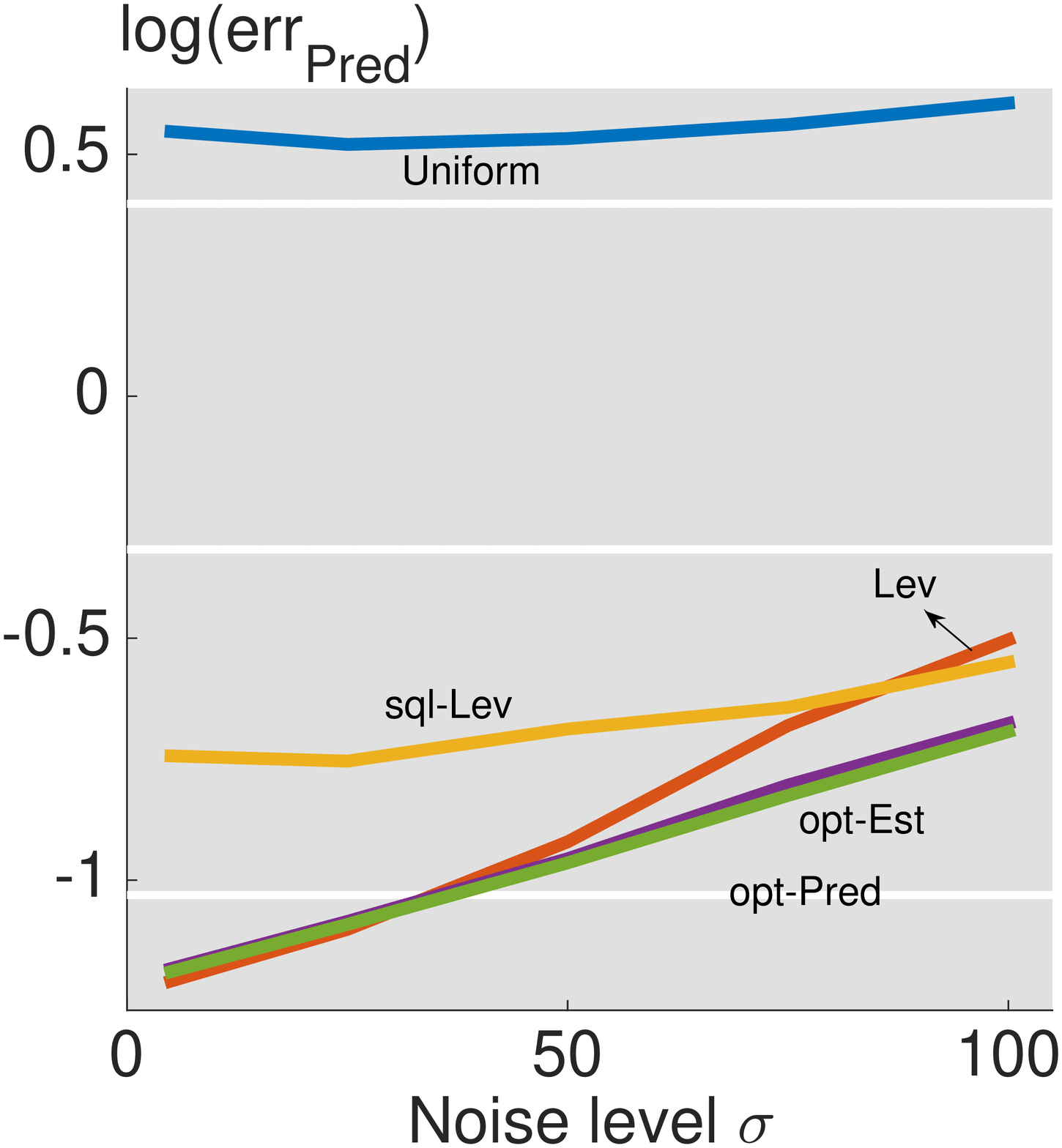} 
\\
 {\small (a) Estimation error .} & {\small (b) Prediction error.} 
 \\ 
\end{tabular}
  \end{center}
   \caption{\label{fig:t_t1_est} Comparison of estimation error  and prediction error of SampleProj as a function of noise level.}
\end{figure}

To study how noise influences the results, given 200 samples, we show the estimation errors and the prediction errors of SampleProj  as a function of noise level. Figure~\ref{fig:t_t1_est} (a) shows that sampling scores for the estimator~\eqref{eq:oss_esti} consistently outperform the other scores. Leverage scores are better than square root of the leverage scores when the noise level is small, and square root of the leverage scores catch up with leverage scores when the noise level increases, which matches Corollary~\ref{cor:lev_esti}.  Figure~\ref{fig:t_t1_est} (b) shows that sampling scores for the predictor~\eqref{eq:oss_pred} consistently outperform the other scores in terms of prediction error; leverage scores are better than square root of the leverage scores when the noise level is small, and square root of the leverage scores catch up with leverage scores when the noise level increases, which matches Corollary~\ref{cor:lev_pred}. 

Now we can answer the two questions proposed in Section~\ref{sec:setup}. Based on SampleProj,
the proposed sampling scores for the estimator consistently outperforms the other sampling scores in terms of the estimation error and the proposed sampling scores for the predictor consistently outperforms the other sampling scores in terms of the prediction error. Leverage score based sampling is better in the noiseless case, but the square roots of the leverage score based sampling is better in the noisy case. Note that many work in computer science theory typically does not address noise and mostly focuses on least squares~\cite{DrineasMM:06, BoutsidisD:09, AvronMT:10}, hence, from the algorithmic perspective, the leverage score sampling was considered optimal.

In the experiments, we use the true NSRs for optimal sampling scores for the estimator and the predictor, which is impractical. In practice, for low noise cases, we set the NSR to zero and sample proportional to $ \sqrt{ \left( \X (\X^T \X)^{-2}  \X^T \right)_{l,l}  (\X \X^T)_{l,l} }$ for the estimator and  $\sqrt{ \Hm_{l,l} (\X \X^T)_{l,l} }$ for the predictor; for high noise cases, we should sample proportional to $ \sqrt{ \left( \X (\X^T \X)^{-2} \X^T \right)_{l,l}  }$ for the estimator and $\sqrt{ \Hm_{l,l} }$ for the predictor. In general, when $\X$ is non-orthonormal, these sampling scores are different than the leverage scores and the square root of the leverage scores.

\section{Conclusions}
We consider the problem of learning a linear model from noisy samples from a statistical perspective. Existing work on sampling for least-squares approximation has focused on using leverage-based scores as an importance sampling distribution. However, it is hard to obtain the precise MSE of the sampled least squares estimator.  To understand the importance sampling distributions, we propose a simple yet effective estimator, called SampleProj, to evaluate the statistical properties of sampling scores. The proposed SampleProj is appealing for theoretical analysis and multitask linear regression. The main contributions are (1) we derive the exact MSE of SampleProj with a given sampling distribution; and (3) we derive the optimal sampling scores for the estimator and the predictor, and show that they are influenced by the noise-to-signal ratio. The numerical simulations show that empirical performance is consistent with the proposed theory. We have derived the optimal sampling strategy for a specific estimator, but identifying the optimal estimator with a computationally efficient sampling strategy remains an open direction.

\bibliographystyle{IEEEbib}
\bibliography{bibl_jelena}

\begin{thebibliography}{10}

\bibitem{ChenVSK:15g}
S.~Chen, R.~Varma, A.~Singh, and J.~Kova{\v c}evi{\'c},
\newblock ``A statistical perspective of sampling scores for linear
  regression,''
\newblock in {\em arXiv:1507.05870}, 2015.

\bibitem{Candes:06}
E.~J. Cand{\`e}s,
\newblock ``{Compressive sampling},''
\newblock in {\em Int. Congr. Mathematicians}, Madrid, Spain, 2006.

\bibitem{Donoho:06}
D.~L. Donoho,
\newblock ``Compressed sensing,''
\newblock {\em IEEE Trans. Inf. Theory}, vol. 52, no. 4, pp. 1289--1306, Apr.
  2006.

\bibitem{TroppLDRB:10}
J.~Tropp, J.~N. Laska, M.~F. Duarte, J.~K Romberg, and R.~G. Baraniuk,
\newblock ``Beyond nyquist: Efficient sampling of sparse bandlimited signals,''
\newblock {\em IEEE Trans. Inf. Theory}, vol. 56, pp. 520 -- 544, June 2010.

\bibitem{HauptCN:11}
J.~Haupt, R.~Castro, and R.~Nowak,
\newblock ``Distilled sensing: Adaptive sampling for sparse detection and
  estimation,''
\newblock {\em IEEE Trans. Inf. Theory}, vol. 57, pp. 6222--6235, Sept. 2011.

\bibitem{DavenportMNW:15}
M.~A. Davenport, A.~K. Massimino, D.~Needell, and T.~Woolf,
\newblock ``Constrained adaptive sensing,''
\newblock {\em arXiv:1506.05889}, 2015.

\bibitem{KrishnamurthyS:13}
A.~Krishnamurthy and A.~Singh,
\newblock ``Low-rank matrix and tensor completion via adaptive sampling,''
\newblock in {\em Proc. Neural Information Process. Syst.}, Harrahs and
  Harveys, Lake Tahoe, Dec. 2013.

\bibitem{PatelGDMB:15}
R.~Patel, T.~A. Goldstein, E.~L. Dyer, A.~Mirhoseini, and R.~G. Baraniuk,
\newblock ``oasis: Adaptive column sampling for kernel matrix approximation,''
  arXiv:1505.05208 [stat.ML], May 2015.

\bibitem{WangS:15a}
Y.~Wang and A.~Singh,
\newblock ``Column subset selection with missing data via active sampling,''
\newblock in {\em AISTATS}, 2015, pp. 1033--1041.

\bibitem{DrineasMM:06}
P.~Drineas, M.~W. Mahoney, and S.~Muthukrishnan,
\newblock ``Sampling algortihms for $l_2$ regression and applications,''
\newblock {\em In Proceedings of the 17th Annual ACM-SIAM Symposium on Discrete
  Algorithms}, pp. 1127--1136, 2006.

\bibitem{ClarksonDMMMW:13}
K.~L. Clarkson, P.~Drineas, M.~Magdon-Ismail, M.~W. Mahoney, X.~Meng, and D.~P.
  Woodruff,
\newblock ``The fast cauchy transform and faster robust linear regression,''
\newblock in {\em Proceedings of the Twenty-Fourth Annual ACM-SIAM Symposium on
  Discrete Algorithms}. SIAM, 2013, pp. 466--477.

\bibitem{MahoneyD:09}
M.~W. Mahoney and P.~Drineas,
\newblock ``Cur matrix decompositions for improved data analysis,''
\newblock {\em Proceedings of the National Academy of Sciences}, vol. 106, no.
  3, pp. 697--702, 2009.

\bibitem{DrineasMMW:12}
P.~Drineas, M.~Magdon-Ismail, M.~W. Mahoney, and D.~Woodruff,
\newblock ``Fast approximation of matrix coherence and statistical leverage,''
\newblock {\em Journal of Machine Learning Research}, vol. 13, no. 1, pp.
  3475--3506, 2012.

\bibitem{AvronMT:10}
H.~Avron, P.~Maymounkov, and S.~Toledo,
\newblock ``Blendenpik: Supercharging lapack's least-squares solver,''
\newblock {\em SIAM Journal on Scientific Computing}, vol. 32, no. 3, pp.
  1217--1236, 2010.

\bibitem{AilonC:10}
N.~Ailon and B.~Chazelle,
\newblock ``Faster dimension reduction,''
\newblock {\em Communications of the ACM}, vol. 53, no. 2, pp. 97--104, 2010.

\bibitem{DrineasMMS:11}
P.~Drineas, M.~W. Mahoney, S.~Muthukrishnan, and T.~Sarl{\'o}s,
\newblock ``Faster least squares approximation,''
\newblock {\em Numerische Mathematik}, vol. 117, no. 2, pp. 219--249, 2011.

\bibitem{MengSM:14}
X.~Meng, M.~A. Saunders, and M.~W. Mahoney,
\newblock ``Lsrn: A parallel iterative solver for strongly over-or
  underdetermined systems,''
\newblock {\em SIAM Journal on Scientific Computing}, vol. 36, no. 2, pp.
  C95--C118, 2014.

\bibitem{Mahoney:11}
M.~W. Mahoney,
\newblock {\em Randomized Algorithms for Matrices and Data},
\newblock Number~2. Now Publishers Inc., Boston, 2011.

\bibitem{WangS:15b}
Y.~Wang and A.~Singh,
\newblock ``An empirical comparison of sampling techniques for matrix column
  subset selection,''
\newblock in {\em Allerton}, 2015.

\bibitem{YangZJZ:15}
T.~Yang, L.~Zhang, R.~Jin, and S.~Zhu,
\newblock ``An explicit sampling dependent spectral error bound for column
  subset selection,''
\newblock {\em arXiv preprint arXiv:1505.00526}, 2015.

\bibitem{Pukelsheim:06}
F.~Pukelsheim,
\newblock {\em Optimal Design of Experiments},
\newblock Classics in Applied Mathematics. SIAM, New York, NY, 2006.

\bibitem{MaMY:15}
P.~Ma, M.~W. Mahoney, and B.~Yu,
\newblock ``A statistical perspective on algorithmic leveraging,''
\newblock {\em Journal of Machine Learning Research}, vol. 16, pp. 861--911,
  2015.

\bibitem{ZhuMMY:15}
R.~Zhu, P.~Ma, M.~W. Mahoney, and B.~Yu,
\newblock ``Optimal subsampling approaches for large sample linear
  regression,''
\newblock {\em arXiv:1509.05111}, 2015.

\bibitem{ChenVSK:15a}
S.~Chen, R.~Varma, A.~Singh, and J.~Kova{\v c}evi{\'c},
\newblock ``Signal recovery on graphs: Random versus experimentally designed
  sampling,''
\newblock in {\em SampTA}, Washington, D.C., 2015.

\bibitem{ChenSMK:14}
S.~Chen, A.~Sandryhaila, J.~M.~F. Moura, and J.~Kova{\v c}evi{\'c},
\newblock ``Signal recovery on graphs: Variation minimization,''
\newblock {\em IEEE Trans. Signal Process.}, vol. 63, pp. 4609--4624, June
  2015.

\bibitem{WengNI:13}
Y.~Weng, R.~Negi, and M.~D. Ili{\'c},
\newblock ``Graphical model for state estimation in electric power systems,''
\newblock in {\em IEEE SmartGridComm Symposium}, Vancouver, Oct. 2013.

\bibitem{HoaglinW:78}
D.~C. Hoaglin and R.~E. Welsch,
\newblock ``The hat matrix in regression and anova,''
\newblock {\em American Statistician}, , no. 1, pp. 17--22, 1978.

\bibitem{VellemanW:81}
P.~F. Velleman and R.~E. Welsch,
\newblock ``Efficient computing of regression diagnotics,''
\newblock {\em American Statistician}, , no. 4, pp. 234--242, 1981.

\bibitem{DhillonLFU:13}
P.~Dhillon, Y.~Lu, D.~P. Foster, and L.~Ungar,
\newblock ``New subsampling algorithms for fast least square regression,''
\newblock in {\em Advances in Neural Information Processing Systems}, 2013, pp.
  360--368.

\bibitem{BoutsidisD:09}
C.~Boutsidis and P.~Drineas,
\newblock ``Random projections for the nonnegative least-squares problem,''
\newblock {\em Linear Algebra Appl.}, vol. 431, pp. 760–771, 2009.

\end{thebibliography}

\appendix
\section{Appendices}
\subsection{Proof of Theorem~\ref{thm:MSE}}
We first show the unbias of the estimator in Lemma~\ref{lem:unbias}. For each element in the bias term, we have
\begin{eqnarray*} 
 && \left( \mathbb{E}  \left[  \widehat{ \beta}  \right] \right)_i  
 \\
& = &  \mathbb{E}_{\M, \epsilon}  \left( \sum_{\M_j \in \M} (\X^{\dagger})_{i,\M_j} \W_{\M_j,\M_j} (\X \beta^{(0)}  + \epsilon)_{\M_j} \right)
\\
& \stackrel{(a)}{=}  & m  \mathbb{E}_{l}  \left( (\X^{\dagger})_{i,l} \frac{1}{m \pi_{l}} (\X \beta^{(0)})_l \right)
\\
& = & m  \sum_{l = 1}^n  (\X^{\dagger})_{i,l} \frac{1}{m \pi_{l}} (\X \beta^{(0)})_l  \pi_l
\\
& = &  \sum_{l =1}^n  (\X^{\dagger})_{i, l}  (\X \beta^{(0)} )_l
\ = \ \left( \beta_1^{(0)} \right)_i,
\end{eqnarray*}
where $(a)$ follows from that each sample is identically and independently draw from $\{\pi_i\}_{i=1}^n$.

We then show the  covariance of the estimator. We first split $ \widehat{ \beta} - \mathbb{E}   \left[ \widehat{ \beta} \right] $ into two components, and then bound both one by one.
\begin{eqnarray*} 
& & \left(  \widehat{ \beta} - \mathbb{E}   \left[ \widehat{ \beta}  \right]   \right)_i
\\
 & = &    \sum_{\M_j \in \M} (\X^{\dagger})_{i,\M_j} \W_{\M_j,\M_j} (\X \beta^{(0)}  + \epsilon)_{\M_j}   -  \left(  \beta^{(0)} \right)_i
   \\
    & = &  \left(  \sum_{\M_j \in \M} (\X^{\dagger})_{i,\M_j} \W_{\M_j,\M_j}  ( \X \beta^{(0)} )_{\M_j} -  ( \beta^{(0)}_1)_i \right)  
    \\
   && ~ +   \sum_{\M_j \in \M} (\X^{\dagger})_{i,\M_j} \W_{\M_j,\M_j}  \epsilon_{\M_j}
    \\
  & = & \Delta^{(1)}_i + \Delta^{(2)}_i.
 \end{eqnarray*}

The first term captures variability due to sampling, while the second term captures variability introduced by noise. Since the noise is independent from the sampling set $\M$, we can bound $\Delta^{(1)}$ and $\Delta^{(2)}$ separately.  For $\Delta^{(1)}_i $, we have
   \begin{eqnarray*} 
 & & \mathbb{E} \left\|   \Delta_i^{(1)}  \right\|^2
 \\
 & = &   \mathbb{E}_{\M}
 \bigg(  \sum_{\M_j \neq \M_{j'} } (\X^{\dagger})_{i,\M_j} (\X^{\dagger})_{i,\M_{j'}}   \W_{\M_j,\M_j} \W_{\M_{j'},\M_{j'}} 
\\ 
&& 
 ( \X \beta^{(0)} )_{\M_j} ( \X \beta^{(0)} )_{\M_{j'}} \bigg) 
+ \mathbb{E}_{\M}
 \bigg(\sum_{\M_j = \M_{j'} } (\X^{\dagger})_{i,\M_j}
  \\
 && 
  (\X^{\dagger})_{i,\M_{j'}}  \W_{\M_j,\M_j} \W_{\M_{j'},\M_{j'}} 
  \\
 &&  
   ( \X \beta^{(0)} )_{\M_j} ( \X \beta^{(0)} )_{\M_{j'}} \bigg) -  ( \beta^{(0)}_1)_i^2 
\end{eqnarray*}
\begin{eqnarray*}
   & = &    \mathbb{E}_{l, l'}
 \bigg( (\X^{\dagger})_{i,l} (\X^{\dagger})_{i,l'} \frac{ m^2-m }{m^2 \pi_l \pi_{l'} } ( \X \beta^{(0)} )_{l}  ( \X \beta^{(0)} )_{l'} \bigg) 
\\
&& 
  + ~
 m \mathbb{E}_{l}
 \left(  (\X^{\dagger})_{i,l}^2 \frac{1}{m^2 \pi_l^2 }   ( \X \beta^{(0)} )_l^2  \right)   -  ( \beta^{(0)}_1)_i^2
 \\
   & = &   \sum_{l,l' = 1}^n (\X^{\dagger})_{i,l} (\X^{\dagger})_{i,l'} \frac{m^2-m}{m^2 \pi_l \pi_{l'} }  ( \X \beta^{(0)} )_{l} ( \X \beta^{(0)} )_{l'} \pi_l \pi_{l'}
\\
 &&  + ~
 m \sum_{l=1}^n
(\X^{\dagger})_{i,l}^2 \frac{1}{m^2 \pi_l^2 }   ( \X \beta^{(0)} )_l^2 \pi_l -  ( \beta^{(0)}_1)_i^2
 \\
   & = & 
 \sum_{l=1}^n  \frac{1}{ m \pi_l }  
(\X^{\dagger})_{i,l}^2 ( \X \beta^{(0)} )_l^2 - \frac{1}{m} ( \beta^{(0)}_1)_i^2.
 \end{eqnarray*}
 
   For the noise term $\Delta^{(2)}_i $, we have
   \begin{eqnarray*} 
 &&  \mathbb{E} || \Delta^{(2)}_i ||^2  
 \ = \  \mathbb{E}_{\M, \epsilon}  \left( \sum_{\M_j \in \M} (\X^{\dagger})_{i,\M_j} \W_{\M_j,\M_j}  \epsilon_{\M_j}  \right) 
 \\
 &&   \bigg( \sum_{\M_{j'} \in \M} (\X^{\dagger})_{i,\M_{j'}}  \W_{\M_{j'},\M_{j'}}  \epsilon_{\M_{j'}}  \bigg) 
\\
& = &  \mathbb{E}_{\M, \epsilon}  \bigg( \sum_{\M_j, \M_{j'} \in \M} (\X^{\dagger})_{i,\M_j} (\X^{\dagger})_{i,\M_{j'}}
\\
&&  \W_{\M_j,\M_j} \W_{\M_{j'},\M_{j'}}  \epsilon_{\M_j} \epsilon_{\M_{j'}} \bigg) 
\\
& =  &  m \mathbb{E}_{l, \epsilon}  \left(  (\X^{\dagger})_{i,l}^2 \frac{1}{m^2 \pi_{l}^2 } \epsilon_l^2 \right) 
\\
& =  &  m  \sum_{l=1}^n  (\X^{\dagger})_{i,l}^2 \frac{1}{m^2 \pi_{l}^2 } \mathbb{E} \left[ \epsilon_l^2 \right] \pi_l
\\
& = & \sigma^2 \sum_{l=1}^n \frac{1}{m \pi_{l} } (\X^{\dagger})_{i,l}^2.
 \end{eqnarray*}

We then combine both $\Delta^{(1)}_i $ and $\Delta^{(2)}_i $, and obtain the variance term.
   \begin{eqnarray*} 
& &  \mathbb{E}  \left\| \widehat{ \beta} - \mathbb{E}   \left[ \widehat{ \beta} \right]   \right\|^2  
\ = \    \sum_{i=1}^{p}   \mathbb{E}    \left(  \widehat{ \beta} - \mathbb{E}   \left[ \widehat{ \beta}  \right]   \right)_i^2
 \\ 
 & = & \sum_{i=1}^{p}   \left(  \mathbb{E} \left\|   \Delta_i^{(1)}  \right\|^2 +  \mathbb{E} \left\|   \Delta_i^{(2)}  \right\|^2 \right)
  \\ 
   & = &
\sum_{i=1}^{p}  \left( \sum_{l=1}^n  \frac{1}{ m \pi_l }  
(\X^{\dagger})_{i,l}^2 \left( ( \X \beta^{(0)} )_l^2 +  \sigma^2 \right) - \frac{1}{m} ( \beta^{(0)})_i^2 \right).
  \end{eqnarray*}

Finally, we put the bias term and the variance term together to obtain the exact MSE of the estimator.
\begin{eqnarray*}
&& \mathbb{E} \left\| \widehat{ \beta } -  \beta^{(0)} \right\|^2
\\
& = &
\mathbb{E} \left\| \widehat{ \beta} -  \mathbb{E}   \left[ \widehat{ \beta } \right] +  \mathbb{E}   \left[ \widehat{ \beta } \right] -  \beta^{(0)}  \right\|^2
\\
& = & \left\| \mathbb{E}   \left[ \widehat{ \beta} \right] -  \beta^{(0)} \right\|^2 +  \mathbb{E} \left\| \widehat{ \beta} -  \mathbb{E}   \left[ \widehat{ \beta} \right]  \right\|^2
\\
& = & 0 +  \sum_{i=1}^{p}  \left( \sum_{l=1}^n  \frac{1}{ m \pi_l }  
(\X^{\dagger})_{i,l}^2 \left( ( \X \beta^{(0)} )_l^2 +  \sigma^2 \right) - \frac{( \beta^{(0)} )_i^2}{m}  \right)
\\
& = &  \sum_{l=1}^n  \frac{ ( \X \beta^{(0)} )_l^2 +  \sigma^2 }{ m \pi_l }   \sum_{i=1}^{p}
(\X^{\dagger})_{i,l}^2 - \frac{  1}{m} \left\|  \beta^{(0)}_1 \right\|_2^2
\\
& = &  {\rm Tr} \left( \X^{\dagger} \W_{\rm C}  (\X^{\dagger})^T \right) -  \frac{  1}{m} \left\|  \beta^{(0)} \right\|_2^2.
\end{eqnarray*}
\hfill$\blacksquare$

\subsection{Proof of Theorem~\ref{thm:MSE}}

To obtain the optimal sampling scores for the estimator, we solve the following optimization problem.
\begin{eqnarray*}
&&  \min_{\pi_l} {\rm Tr} \left( \X^{\dagger} \W_{\rm C}  (\X^{\dagger})^T \right),  
 \\
&& {\rm subject~to:}~ \sum_l  \pi_l = 1, \pi_l \geq 0.
\end{eqnarray*}
The objective function is the upper bound on the MSE of the estimator derived in Theorem~\ref{thm:MSE} and the constraints require $\{ \pi_l \}_{i=1}^n$ to be a valid probability distribution. The Lagrangian function is then
\begin{eqnarray*}
L( \pi_l, \lambda, \mu_l  ) & = &   \sum_{l=1}^n  \frac{ (\X \beta^{(0)})_l^2 + \sigma^2 }{ m \pi_l }  
\sum_{i=1}^{p} (\X^{\dagger})_{i,l}^2 
\\
&& + \lambda \left( \sum_l  \pi_l - 1 \right) + \sum_l \mu_l \pi_l.
\end{eqnarray*}

We set the derivative of the Lagrangian function to zero,
\begin{eqnarray*}
 \frac{d L}{d \pi_l} & = &  -  \frac{  (\X \beta^{(0)})_l^2 + \sigma^2 }{ m \pi_l^2 }  
\sum_{i=1}^{p} (\X^{\dagger})_{i,l}^2 + \lambda + \mu_l = 0,
 \end{eqnarray*}
 and then, we obtain the optimal sampling distribution to be
 \begin{eqnarray*}
  \pi_l  & \propto & \sqrt{  \left( \sum_{i=1}^{p}(\X^{\dagger})_{i,l}^2 \right)  \left(  (\X \beta^{(0)})_l^2 + \sigma^2 \right)  }
  \\
  & = &  \sqrt{ \left( (\X^{\dagger})^T \X^{\dagger} \right)_{l,l}  \left( (\X \beta^{(0)})_l^2  + \sigma^2 \right) }.
\end{eqnarray*}

Similarly,  we minimize ${\rm Tr} \left( \Hm \W_{\rm C} \right)$ to obtain the optimal sampling distribution for the predictor to be
\begin{eqnarray*}
 \pi_l \propto \sqrt{ \Hm_{l,l} \left( (\X \beta^{(0)})_l^2 + \sigma^2 \right) }.
\end{eqnarray*}
\hfill$\blacksquare$

\end{document}